\documentclass[10pt,twocolumn,letterpaper]{article}

\usepackage{cvpr}              

\usepackage{graphicx}
\usepackage{amsmath}
\usepackage{amssymb}
\usepackage{booktabs}

\usepackage{tablefootnote}
\usepackage{amsmath}
\usepackage{algorithm}
\usepackage[noend]{algpseudocode}

\usepackage[table]{xcolor}
\colorlet{lightred}{red!20}
\colorlet{lightblue}{blue!20}

\newcolumntype{V}{>{\centering\arraybackslash} m{.4\linewidth} }

\usepackage[pagebackref,breaklinks]{hyperref}

\usepackage[capitalize]{cleveref}
\crefname{section}{Sec.}{Secs.}
\Crefname{section}{Section}{Sections}
\Crefname{table}{Table}{Tables}
\crefname{table}{Tab.}{Tabs.}

\usepackage{microtype}
\usepackage{comment}
\usepackage{balance}

\def\cvprPaperID{46} 
\def\confName{CVPR}
\def\confYear{2022}

\usepackage{xcolor}

\newcommand*{\affaddr}[1]{#1} 
\newcommand*{\affmark}[1][*]{\textsuperscript{#1}}

\begin{document}
\title{TorMentor: Deterministic dynamic-path, data augmentations with fractals}
\author{%
\href{{mailto:anguelos.nicolaou\@gmail.com}}{Anguelos Nicolaou}\affmark[1], 
\href{vincent.christlein@fau.de}{Vincent Christlein}\affmark[2], 
\href{edgar.riba@gmail.com}{Edgar Riba}\affmark[3], \href{sj8716643@126.com}{Jian Shi}\affmark[3], 
\href{georg.vogeler@uni-graz.at}{Georg Vogeler}\affmark[1], \href{mathias.seuret@fau.de}{Mathias Seuret}\affmark[2]\\
\affaddr{\affmark[1]University of Graz}, 
\affaddr{\affmark[2]Friedrich-Alexander-Universit\"at Erlangen-N\"urnberg},
\affaddr{\affmark[3]kornia.org}
%
}




\maketitle

\begin{abstract}
We propose the use of fractals as a means of efficient data augmentation.
Specifically, we employ plasma fractals for adapting global image augmentation transformations into continuous local transforms. 
We formulate the diamond square algorithm as a cascade of simple convolution operations allowing efficient computation of plasma fractals on the GPU.
We present the TorMentor image augmentation framework that is totally modular and deterministic across images and point-clouds.
All image augmentation operations can be combined through pipelining and random branching to form flow networks of arbitrary width and depth.
We demonstrate the efficiency of the proposed approach with experiments on document image segmentation (binarization) with the DIBCO datasets. 
The proposed approach demonstrates superior performance to traditional image augmentation techniques.
Finally, we use extended synthetic binary text images in a self-supervision regiment and outperform the same model when trained with limited data and simple extensions. 
\end{abstract}

\begin{figure}[pb]
\centerline{
\includegraphics[width=.31\columnwidth]{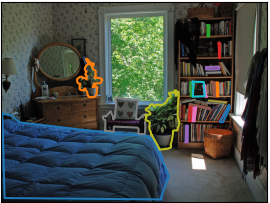}
\includegraphics[width=.31\columnwidth]{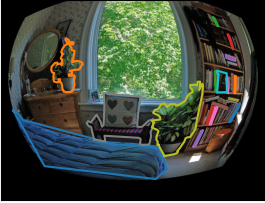}
\includegraphics[width=.31\columnwidth]{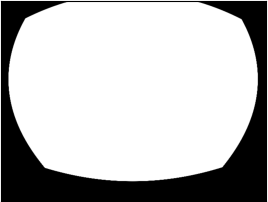}
}
\caption{An MS-COCO~\cite{lin2014microsoft} sample and its ground truth augmented through TorMentor along with the automatically generated validity mask}
\label{fig:tease}
\end{figure}

\section{Introduction}
\label{sec:intro}

In the era of deep learning, computer vision-based neural networks need extreme quantities of annotated data.
This has led to Image Data Augmentation (IDA) being employed in any state-of-the-art training pipeline to various degrees.
Specialised domains such as Document Image Analysis (DIA) or medical imaging usually are much more restricted in the size of the available datasets making data augmentation an integral part of the training pipeline.
When the use case exceeds typical image classification such as image segmentation, object detection, landmark detection etc., data augmentation becomes more challenging as the ground truth needs to be modified along with the input sample.

Image data augmentation is the introduction of noise in data samples, essentially creating new plausible samples in order to train a ML model with more data.
This causes the model to become invariant to the specific kind of noise.
Each noise operation of an IDA tries to mimic distortions that could occur naturally in an image and thus defining and tuning it requires domain-specific knowledge.

IDA operators can also be employed in self-supervision scenarios, test-time augmentation, and ablation studies. 

One could argue that defining all the invariants that need to be learned, is essentially defining the problem itself.


\section{Plasma Fractals for Image Augmentation}
\label{sec:fractals}
Elastic transforms have proved quite important for augmenting data, especially textual images\cite{simard2003best} but they require employing large Gaussian filters.
The diamond-square algorithm was proposed~\cite{fournier1982computer} in 1982 as an algorithm for generating random height-maps for computer graphics.
The algorithm can be used to generate cloud-looking fractals called plasma-fractals.
While the popular algorithm requires sparse memory access, and square images of size $(2^n+1)\times(2^n+1)$ for any $n>2$, we propose a version of the algorithm that can be implemented with convolutions.

    \begin{algorithm}[t]
        \caption{Convolutional Diamond Square}\label{al:odos}
        \label{alg:ds}
        \begin{algorithmic}[1]
            \Procedure{OneDS}{$plasma$, $e$}    
            \State$dfilter\gets [[0.25, 0, 0.25], [0, 0, 0], [0.25, 0, 0.25]]$
            \State$sfilter\gets [[0, 0.25, 0], [0.25, 0, 0.25,], [0, 0.25, 0]]$
            \State$oldw, oldh\gets size(plasma)$
            \State$w, h\gets (oldw-1)*2+1, (oldh-1)*2+1 $
            \State$border\gets ones(w, h)$
            \State$border[1:-1, 1:-1]\gets ones(w-2, h-2)$
            \State$rnd\gets random(w, h)$
            \State$dilated\gets zeros(w,h)$            
            \State$dilated[::2,::2]\gets plasma + random(w, h)*e $
            \State$d \gets dilated \circledast dfilter$
            \State$dc \gets ispositive(d)$
            \State$dilated\gets dilated + (1 -e)* d * dc + rnd*dc* e$
            \State$s \gets dilated \circledast sfilter * border $
            \State$sc \gets ispositive(s)$
            \State \textbf{return} $dilated +(1-e) * d *sc + sc * rnd * e$
            \EndProcedure
            \Procedure{DS}{$steps$,$roughness$}
            \State $e\gets 1$
            \State $plasma\gets rand(3,3)$
            \For{$i\gets 1, steps$}
            	\State $e\gets e * roughness$
            	\State $plasma\gets OneDS(plasma, e)$
            \EndFor
            \State \textbf{return} $plasma$
            \EndProcedure
        \end{algorithmic}
    \end{algorithm}

\begin{figure*}
\centering
\begin{tabular}{cccccccc}
\includegraphics[width=.1\textwidth]{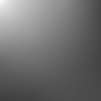} &
\includegraphics[width=.1\textwidth]{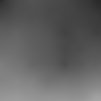} &
\includegraphics[width=.1\textwidth]{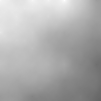} &
\includegraphics[width=.1\textwidth]{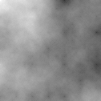} &
\includegraphics[width=.1\textwidth]{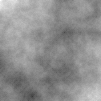} &
\includegraphics[width=.1\textwidth]{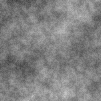} &
\includegraphics[width=.1\textwidth]{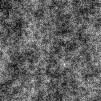} &
\includegraphics[width=.1\textwidth]{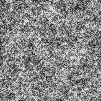} 
\end{tabular}
\caption{Generated plasma fractals $129\times 129$ pixels with a roughness of 0.2 to 0.9 }
\label{fig:plasma}
\end{figure*}

In algorithm~\ref{alg:ds}, the python-numpy inspired pseudo-code description of the proposed algorithm can be seen. 
The algorithm consists of two steps: the first \textit{\textbf{\textsc{OneDS}}} is applying a diamond and square step cascade on an existing image and some weighted random pixels effectively quadrupling its resolution.
The second step \textbf{\textsc{DS}} invokes \textit{\textbf{\textsc{OneDS}}} recursively in order to grow the plasma fractal to an arbitrary size.
It should be pointed out that in line 19, plasma could be initialized to any image of odd dimension sizes larger than 3, but if a dimension is initialized with a size grater than 3, then it will be missing some low frequencies.
The $roughness$ parameter, through $e$, controls the ratio between the existing pixels and random pixels added.
In practice other than the desired resolution (recursion steps), it is the only parameter controlling the fractal generation and can be perceived as a parameter controlling whether low or high frequencies will dominate.
In \cref{fig:plasma}, the resulting plasma fractals for different roughness values can be seen.

\begin{figure}[tb]
\centerline{
\includegraphics[width=.9\columnwidth]{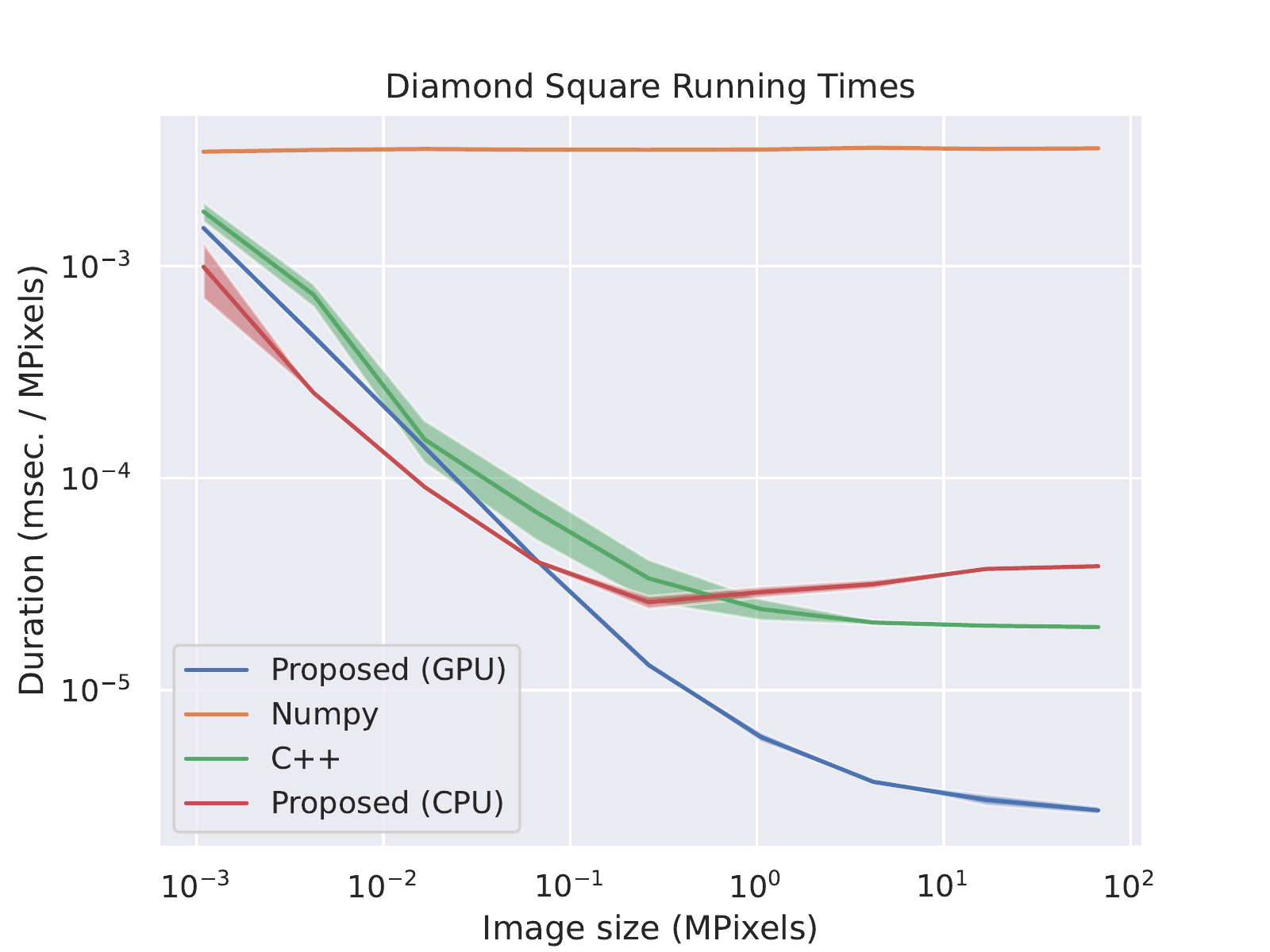}
}
\caption{Proposed Diamond Square Benchmark.}
\label{fig:ds_benchmark}
\end{figure}

Other than the fact that \textit{\textbf{\textsc{OneDS}}} is differentiable with respect to $plasma$ and can be used as a valid neural network layer, the proposed algorithm can be computed efficiently on the GPU with PyTorch~\cite{pytorch_nips}.
In Fig.~\ref{fig:ds_benchmark}, the performance with respect to the output image size is compared between a C++ implementation~\cite{roukien}, a generic python-numpy version~\cite{scipython}, and the proposed method in CPU and GPU mode.
They were tested for resolutions of $65\times65$ up to $8193\times8193$.
The proposed method on CPU (single thread) converges to being twice as slow as the C++ version while the GPU version is more than an order of magnitude faster.

\begin{figure*}[!ht]
\centering
\setlength{\tabcolsep}{1px}
\begin{tabular}{lcccc}
Augmentation & Printed & Handwritten & Synthetic 1 & Synthetic 2 \\ \noalign{\smallskip} 
None &
\raisebox{-.5\height}{
\includegraphics[width=.15\textwidth]{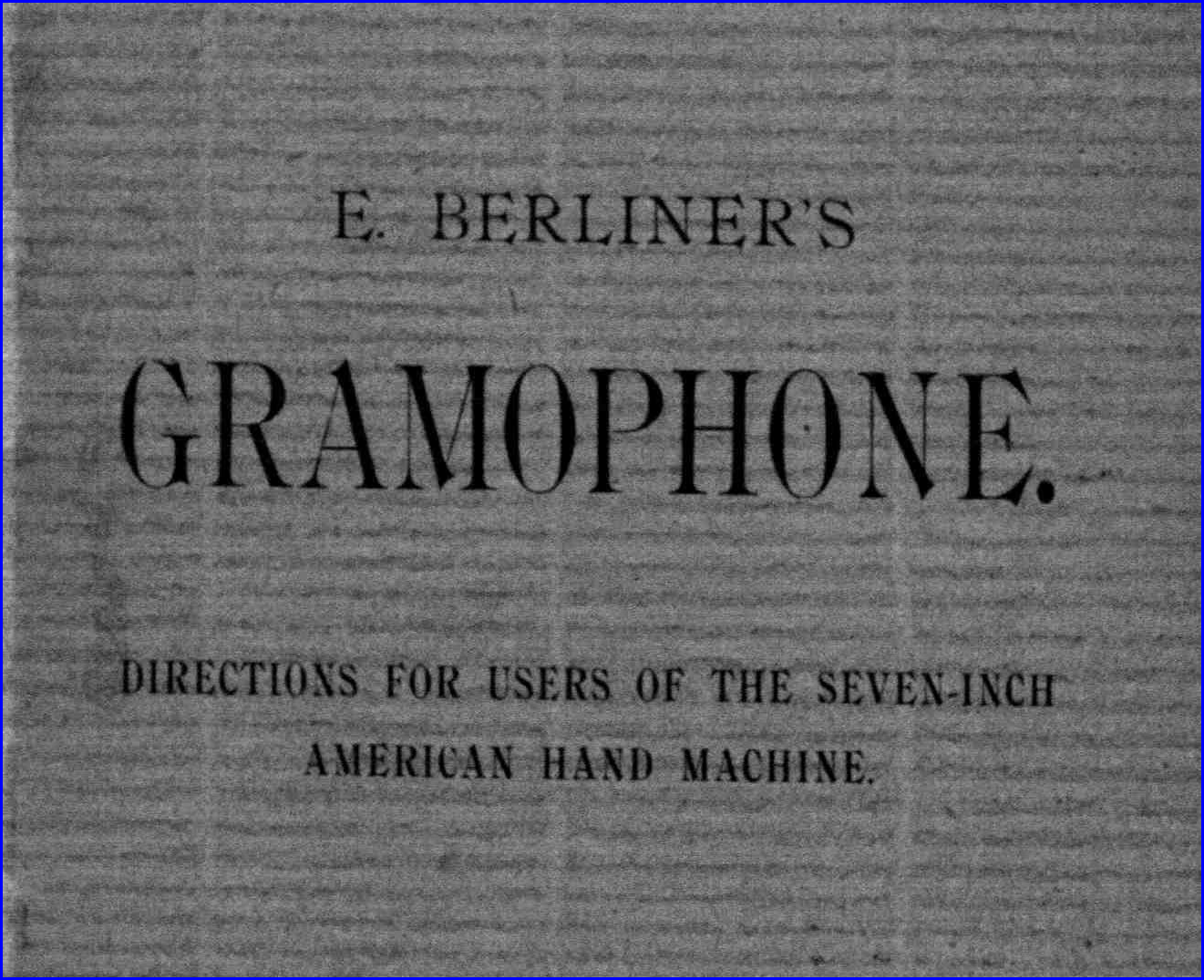}
}
&
\raisebox{-.5\height}{
\includegraphics[width=.145\textwidth]{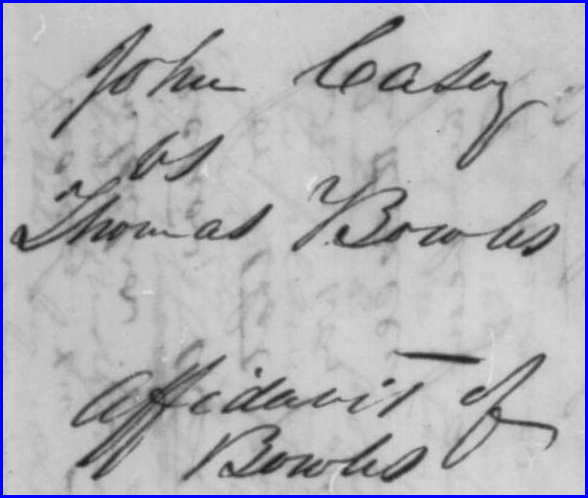} 
}
& 
\raisebox{-.5\height}{
\includegraphics[width=.295\textwidth]{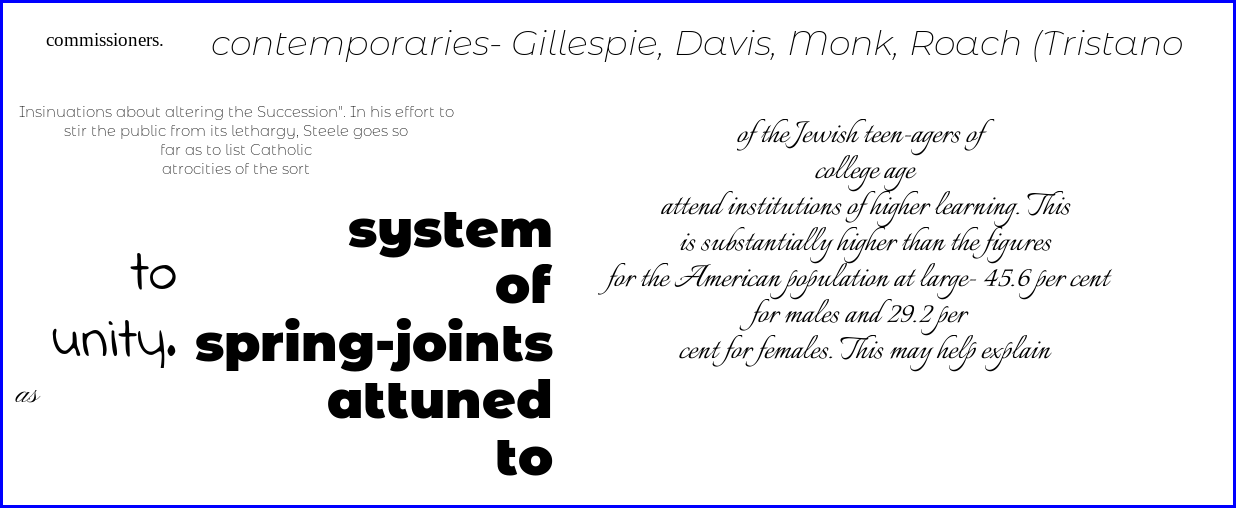} 
}
&
\raisebox{-.5\height}{
\includegraphics[width=.205\textwidth]{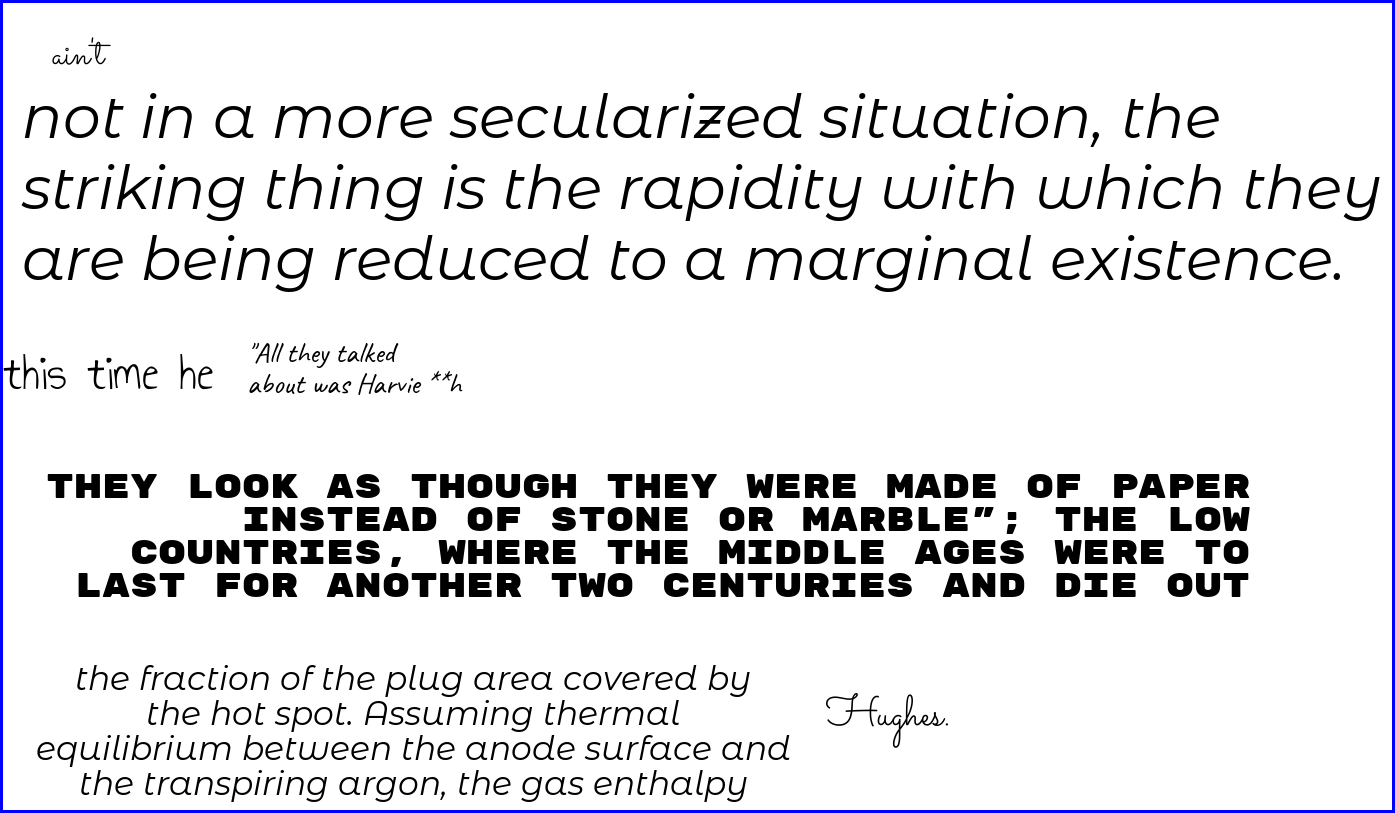}
}
\\  \noalign{\smallskip}
Global  &
\raisebox{-.5\height}{
\includegraphics[width=.15\textwidth]{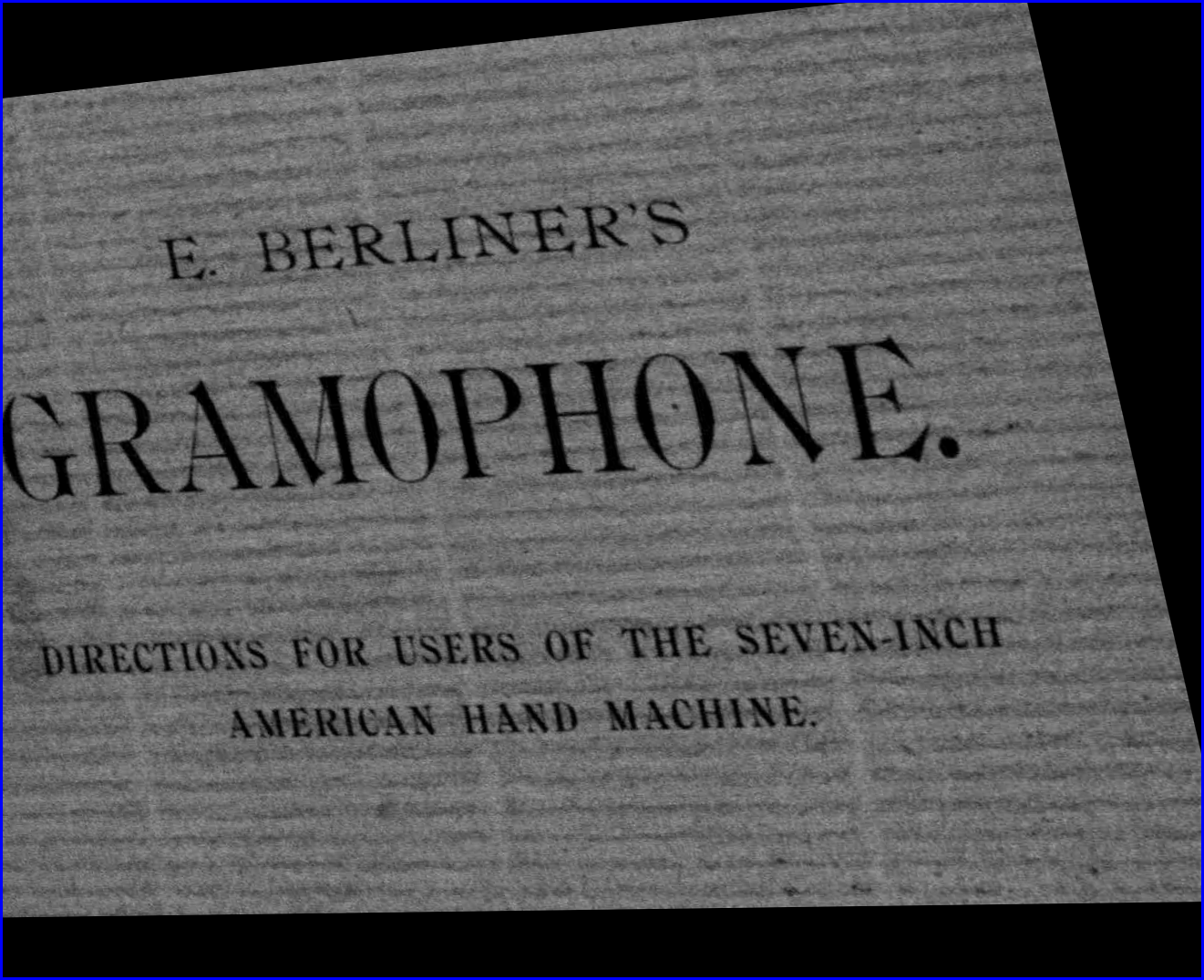}
}
&
\raisebox{-.5\height}{
\includegraphics[width=.145\textwidth]{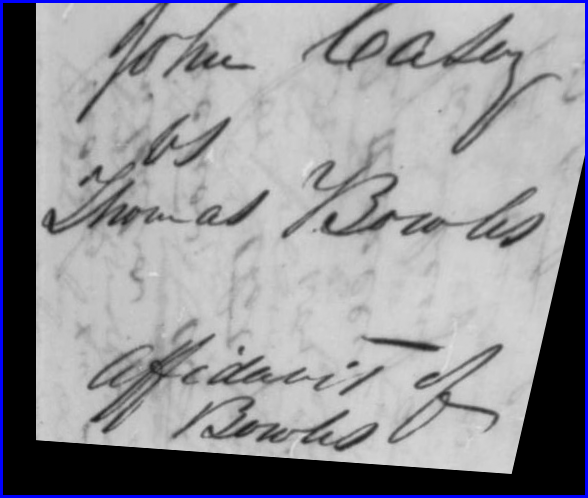}
}
&
\raisebox{-.5\height}{
\includegraphics[width=.295\textwidth]{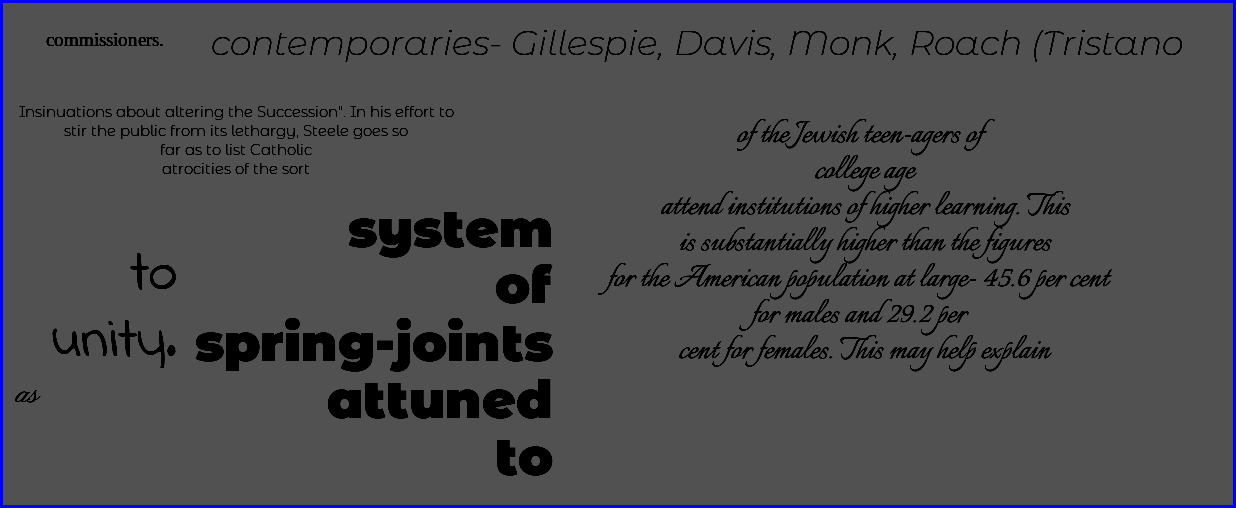}
}
&
\raisebox{-.5\height}{
\includegraphics[width=.205\textwidth]{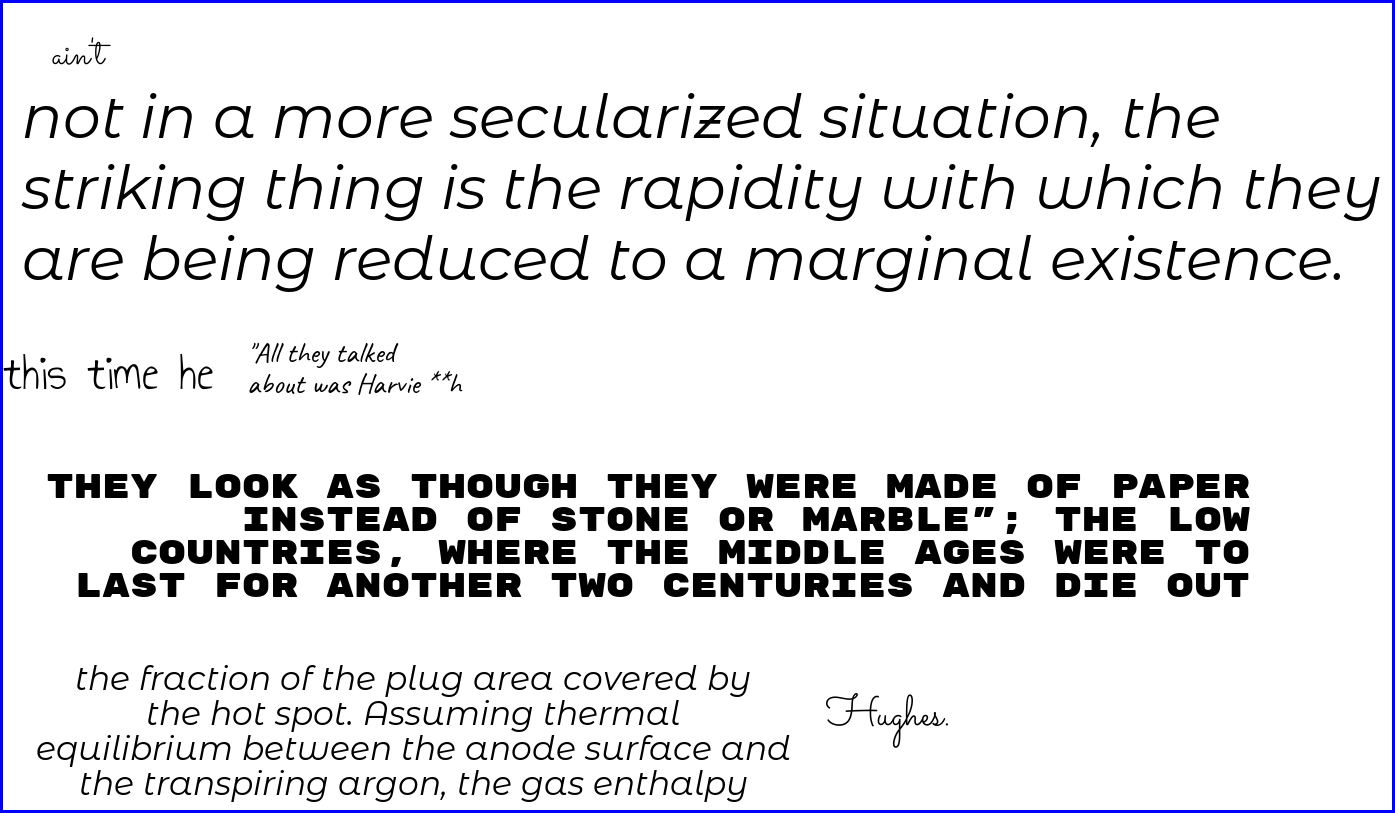}
}
\\  \noalign{\smallskip}
Plasma Cascade 
&
\raisebox{-.5\height}{
\includegraphics[width=.15\textwidth]{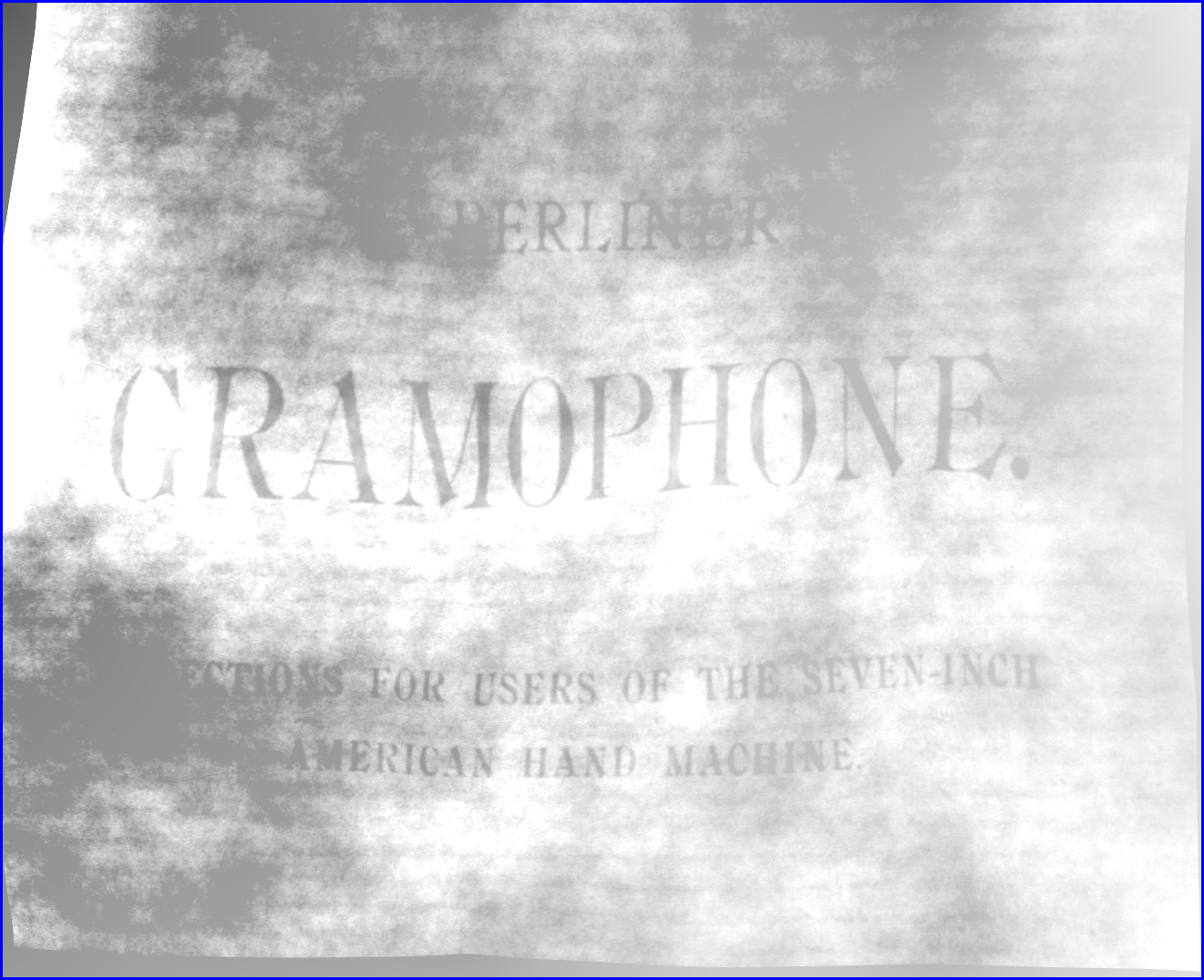} 
}
&
\raisebox{-.5\height}{
\includegraphics[width=.145\textwidth]{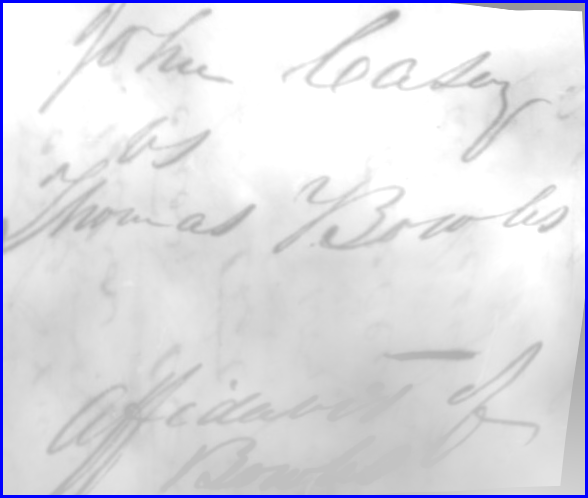} 
}
& 
\raisebox{-.5\height}{
\includegraphics[width=.295\textwidth]{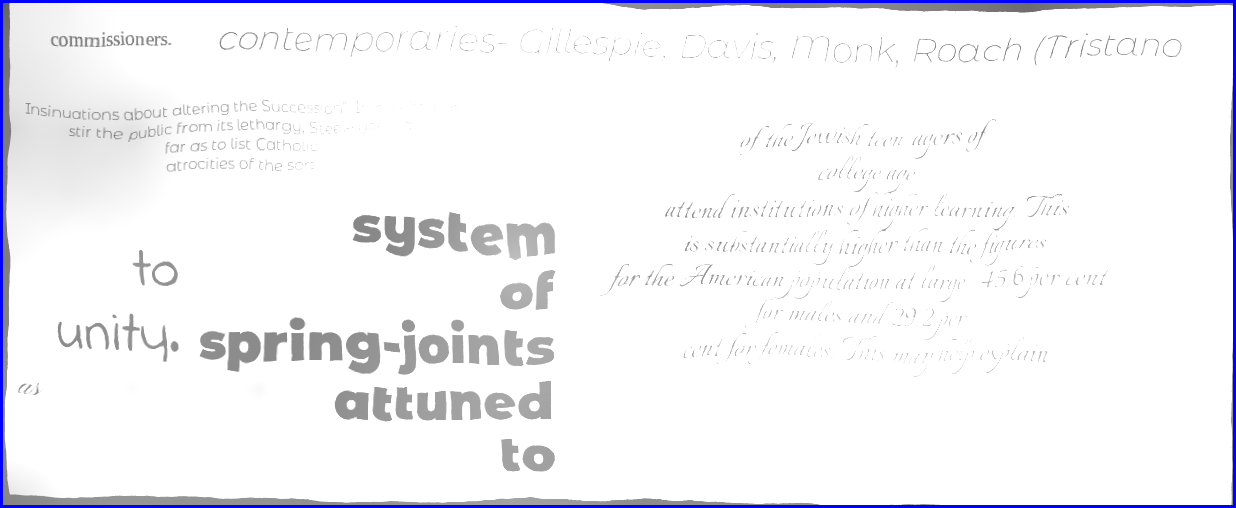} 
}
&
\raisebox{-.5\height}{
\includegraphics[width=.205\textwidth]{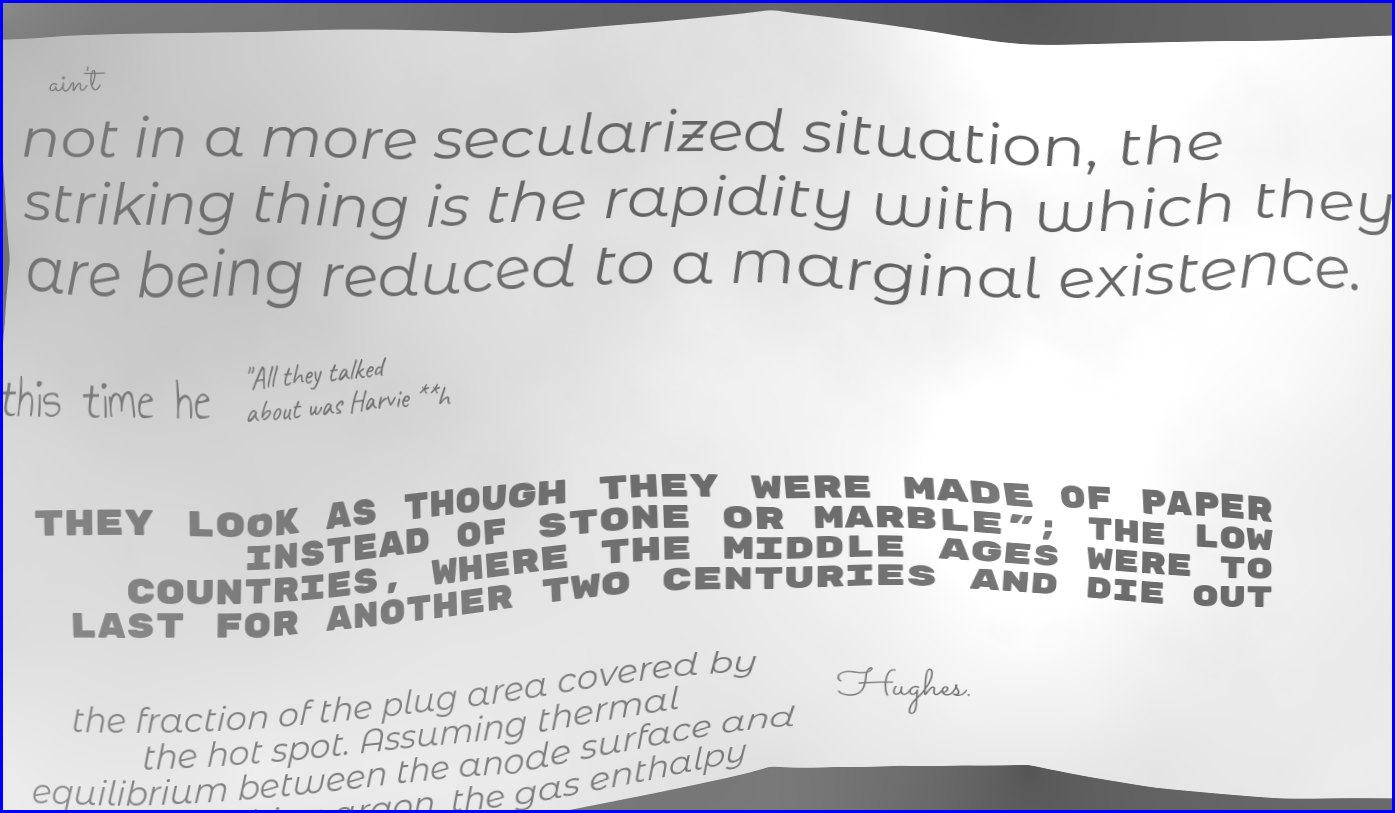}
}
\\ \noalign{\smallskip}
Plasma Branching 
&
\raisebox{-.5\height}{
\includegraphics[width=.15\textwidth]{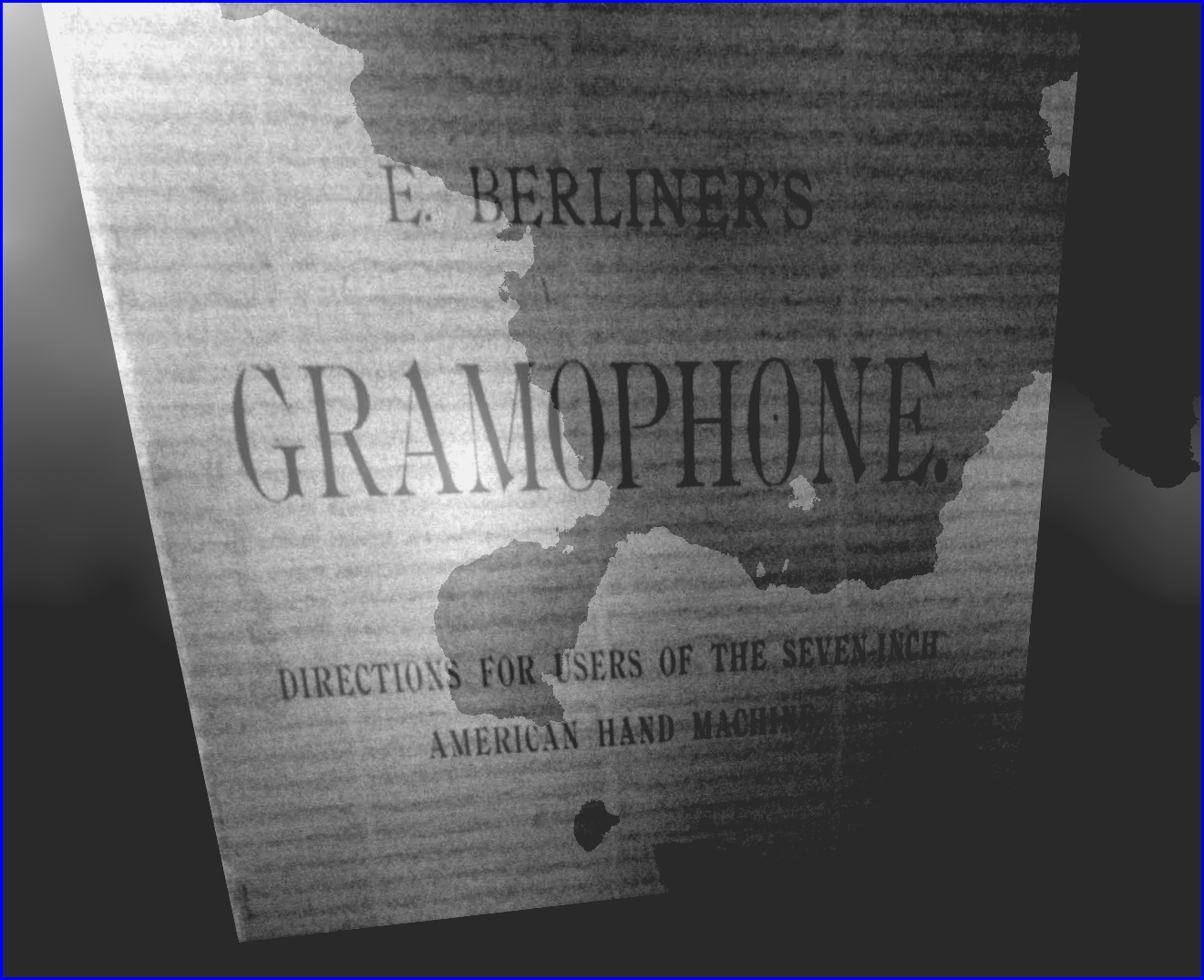} 
}
&
\raisebox{-.5\height}{
\includegraphics[width=.145\textwidth]{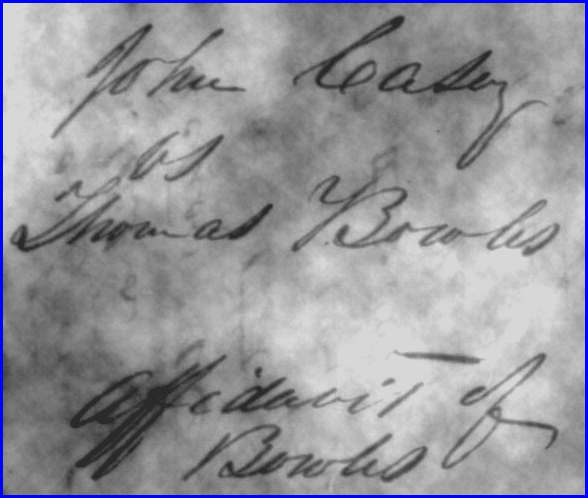} 
}
&
\raisebox{-.5\height}{
\includegraphics[width=.295\textwidth]{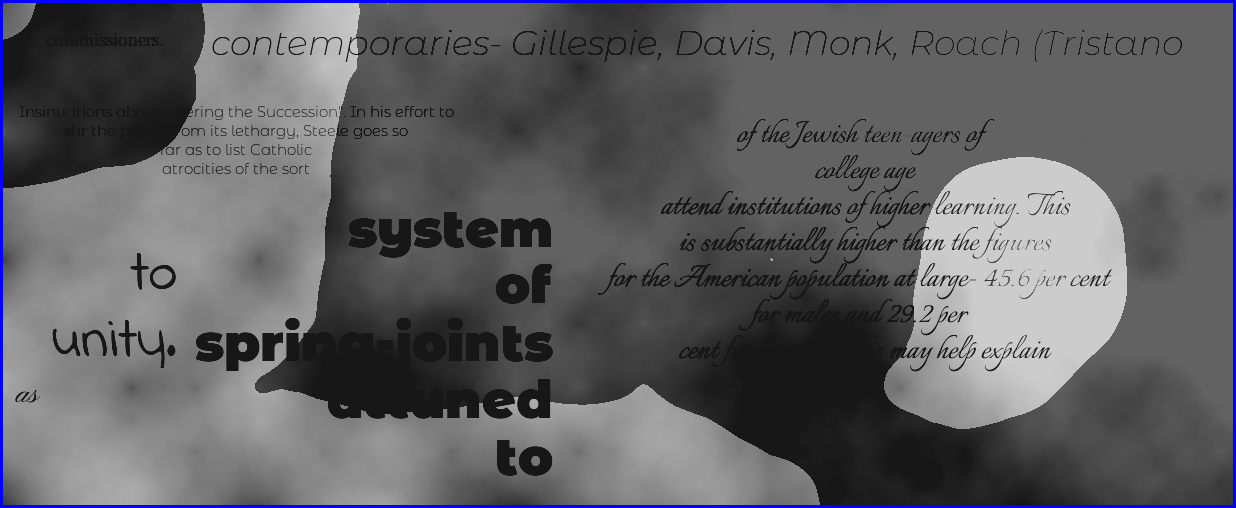} 
}
&
\raisebox{-.5\height}{
\includegraphics[width=.205\textwidth]{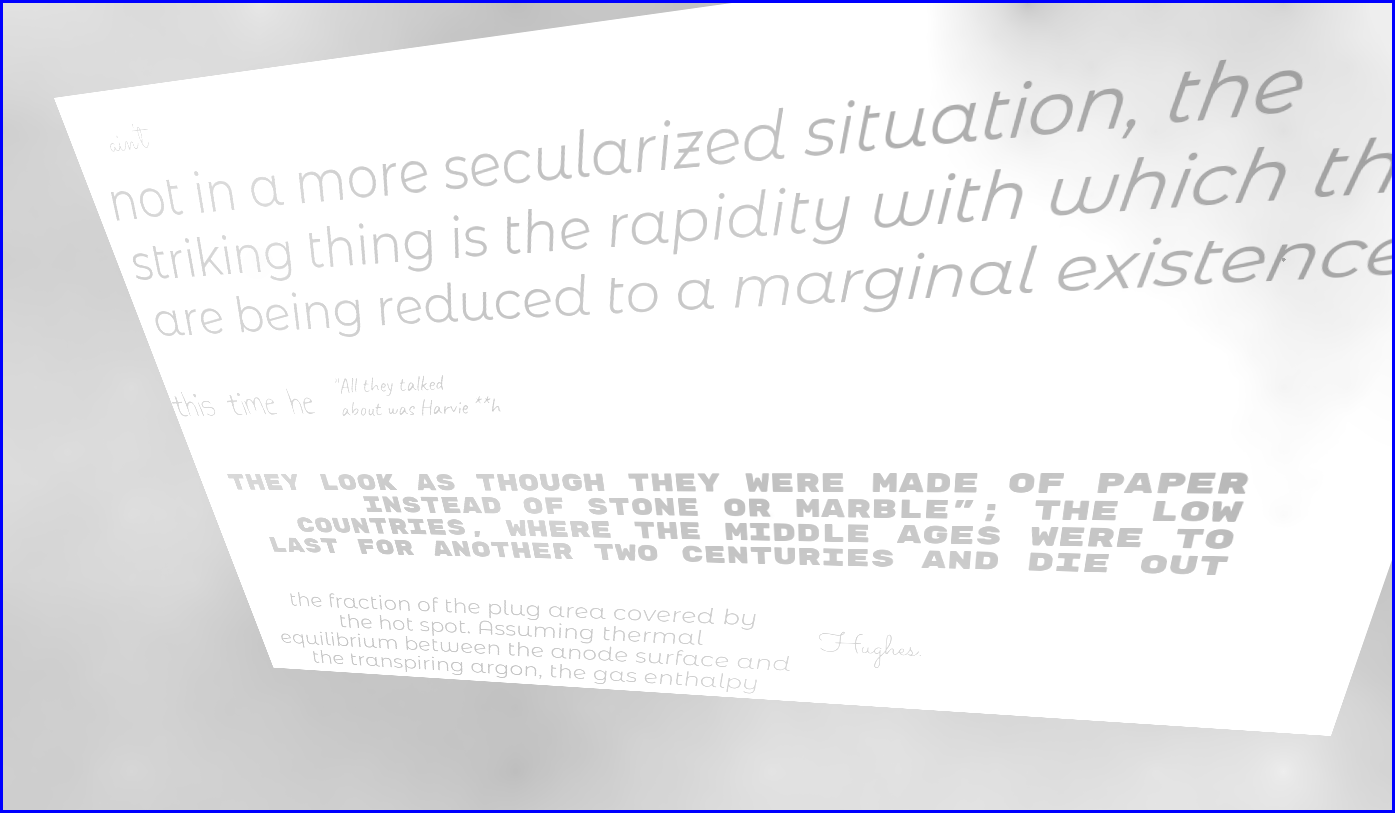} 
}
\\  \noalign{\smallskip}
\end{tabular}
\caption{Samples augmented under the three augmentation regiments in the document segmentation experiment. 
}
\label{fig:samples}
\end{figure*}
\section{The TorMentor augmentation framework}
\label{sec:tormentor}
The extensive use of plasma-fractals to provide localized augmentation operations was coupled with some other design patterns into a coherent image data augmentation framework that called TorMentor.
\footnote{\url{https://github.com/anguelos/tormentor}.}

At the heart of the design lies the concept of ventral operations, which move things in the image, and dorsal augmentations, which modify pixels where they are~\cite{ericsson2021self}.
Ventral operations affect images, masks, and points while dorsal operations affect only images, this allows for the creation of deep augmentation pipelines that are applicable at the same time on image pointclouds and masks.

An augmentation operation is a class that possess the random generation parameters, instances of that class are lightweight objects containing a random number generator seed as their only data member allowing each instance of the operation to be deterministic and serializable.
Each augmentation class must implement a method sampling the random distributions that specify the augmentation parameters given the size of the data and a functional method that applies it on a sampling field if it is a ventral operation or directly on the image if it is a dorsal operation.

Finally, other than ventral and dorsal augmentations, a random choice between several augmentations, a cascade of several augmentations, and an identity augmentation are also defined. 
These preserve determinism for data of the same (image) size and allow for combinations of elementary augmentations into ones of arbitrary complexity, \eg an operation known as random flip, can be implemented in tormentor as a choice between a vertical flip, a horizontal flip, a cascade of both, or an identity.
In essence, a TorMentor's data augmentation regiment defines a flow network where the input data are the source, the augmented data are the sink, and every augmentation instance combined with a specific input image size defines a random path from the source to the sink.

As with every other augmentation class, choices have their own categorical distributions from which they sample and these are easily tuned to control the probability of every branch in the graph.

TorMentor builds on top of Kornia~\cite{riba2020kornia} and thus it is differentiable and a PyTorch layer.
This allows to employ augmentations anywhere inside an end-to-end model training regiment including between the generator and discriminator in a Generative Adversarial Network (GAN).

\begin{table*}[t]
\setlength{\tabcolsep}{4pt}
    \centering
    \caption{Document Image Segmentation Experiment}
    \begin{tabular}{lccccccccccc}
        \toprule
        \multicolumn{2}{c}{} & \multicolumn{9}{c}{FScore \% on DIBCO Dataset} \\ 
        \cmidrule{3-12}
        Method & Augmentation &  2009 & 2010 & 2011 & 2012 & 2013 & 2014 & 2016 & 2017 & 2018 & 2019 \\ 
        \midrule
\rowcolor{white}\multicolumn{2}{l}{DIBCO Participant centile 100\,\%} & 91.24 & 91.78 & 93.33 & 92.85 & 92.7 & 96.88 & 88.72 & 91.04 & 89.37 & 72.88 \\
\rowcolor{white}\multicolumn{2}{l}{DIBCO Participant centile \phantom{0}75\,\%} & 86.17 & 87.98 & 91.68 & 90.035 & 89.78 & 94.17 & 88.14 & 86.38 & 82.91 & 62.76 \\
\rowcolor{white}\multicolumn{2}{l}{DIBCO Participant centile \phantom{0}50\,\%} & 84.57 & 85.06 & 88.99 & 89.38 & 89.06 & 89.51 & 87.61 & 83.10 & 78.46 & 57.66 \\
\midrule
\rowcolor{white}Otsu & -- & 78.60 & 85.43 & 82.10 & 75.07 & 80.03  & 86.59 & 77.74 & 51.46 & 51.455 & 47.83 \\
\rowcolor{white}Thr. Oracle  & -- & 87.86 & 87.70 & 85.88 & 87.81 & 87.61 & 90.66 & 85.15 & 88.74 & 88.741 & 81.13 \\
\midrule
\rowcolor{lightblue}BIUnet & None & -- & 90.52 & 87.86 & \textbf{89.92} & \textbf{90.42} & 91.39 & 84.62 & 87.59 & 66.43 & 62.03 \\ 
%
%
%
\rowcolor{lightblue}BIUnet & Global & -- & 75.91 & 86.07 & 79.50 & 82.76 & 82.05 & 79.71 & 82.75 & 59.81 & 51.35 \\
%
\rowcolor{lightblue}BIUnet & Plasma Cascade & -- & \textbf{91.92} & 85.52 & 88.83 & 89.41 & \textbf{94.03} & 85.00 & 86.30 & \textbf{83.42} & 63.62 \\
%
\rowcolor{lightblue}Synth  BIUnet & Plasma Branching & 87.85 & 87.78 & \textbf{88.45} & 87.35 & 89.34 & 90.39 & \textbf{89.07} & \textbf{89.69} & 82.82 & \textbf{69.84} \\
%
%
        \bottomrule
    \end{tabular}
    \label{tab:dibco}
\end{table*}

\section{Document Image Segmentation Experiments}
\label{sec:experiments}
We performed several experiments for document image segmentation (binarization).
Our intention was not to compare the models we employed to the state of the art but rather obtain insights about generalization abilities of a straight-forward model, trained under different augmentation strategies.
Document Image Segmentation (DIS) ground-truthing is extremely labor intensive and in the case of historical documents, paleographers might be needed.
In most cases, training data are limited to at most a few pages.

We used the whole range of the DIBCO datasets that were publicly available~\cite{dibco2009, dibco2010, dibco2011,dibco2012,dibco2013,dibco2016,dibco2017,dibco2018, dibco2019}.
In all experiments we performed, either we used DIBCO2009 as a train set or no ``real'' data at all. 
We used as a metric for each page, the FScore, the harmonic mean of precision and recall as defined in~\cite{dibco2009}, and averaged it across all samples in each dataset regardless of the sample size.
While DIBCO has different tracks and modalities in several years, we averaged the FScore across all samples in all cases among the dataset.
We also created a synthetic dataset with 30 images containing black text on various fonts and sizes rendered as small text-blocks on white pages; these images are employed in a self supervision task being both input and output but augmenting the input.
While synthetic data has been used for text image classification\cite{jaderberg2014synthetic}, to the authors knowledge, they have not been used for text-image segmentation.
In \cref{fig:samples}, DIBCO and synthetic samples can be seen on the first row while rows 2, 3, and 4 demonstrate how they are augmented through the different augmentation regiments.

We chose a lightweight segmentation model: iUNets~\cite{etmann2020iunets} which are reversible~\cite{vandeLeemput2019MemCNN} UNets~\cite{ronneberger2015u}.
Networks such as the UNet are quite small from the perspective of parameters but the memory they require is vast and proportional to the size of the input image.
Reversible networks allow to economize memory needed for caching the forward pass by an order of magnitude allowing to use the UNet in a fully convolutional way without the use of patches and their associated complications such as stitching and border artifacts.
As the point of the experiments was not to compete with the state of the art in DIS, we chose a light-weigh architecture with 1,597,698 parameters, we also omitted connected component post-processing heuristics that typically improve the method outputs.

We also measured the performance of two global thresholds as reference, Otsu's threshold~\cite{otsu} and the Threshold oracle~\cite{barron2020generalization}.

Finally for reference, we provide quantiles 100\,\% (best), 75\,\%, and 50\,\%(median) from all participating methods as reported by the competition organisers.


All images were converted to grayscale and three augmentation regiments were defined: a traditional one, called \textit{Global}, selecting randomly between a perspective transformation, a brightness modification, and Gaussian additive noise. A cascade of three plasma-based augmentations: a plasma-brightness operation followed by plasma-wrapping, followed by a linear color space manipulation. And a \textit{Branching} regiment where a cascade of several choices between mostly plasma-based augmentations was performed.

The results presented in \cref{tab:dibco} allow several observations.
The \textit{Global} regiment performs worse than no augmentation at all.
The model trained on synthetic images achieves the best performance in 4 out of 9 datasets, compared to competition participants it would rank above or near the top 25\,\% and in recent years where the data became more challenging even higher.

Even when training on 10 images and without augmentation, the BIUnet could not overfit the data as the images were much larger than its receptive field demonstrating the merit of training fully convolutionally instead of patched-based.
The DIBCO datasets are quite different from year to year and demonstrate very well that there is no such thing as the overall best DIS method.

\section{Conclusion Discussion and Future Work}
\label{sec:conclusion}

TorMentor is developed as a technology demonstrator that enables the encoding of domain-specific expert knowledge to generate plausible distortions.
While defining an augmentation is constrained by its formalism, the fact that an augmentation is defined once and automatically applied on images, pointclouds, etc., makes it much easier to maintain than popular frameworks such as Albumentations~\cite{albumentations} that redefine operations for every kind of data.
The choice and cascade augmentations are also created through the \textbf{\textit{`\symbol{94}'}} and \textbf{\textit{`$|$'}} operators respectively allowing for readable and dense pythonic constructs of complicated augmentation graphs.
The fact that it could be used to successfully train a UNet strictly on augmented ground truth maps proves its potential.

\begin{figure}[ht]
\centerline{\includegraphics[width=.75\columnwidth]{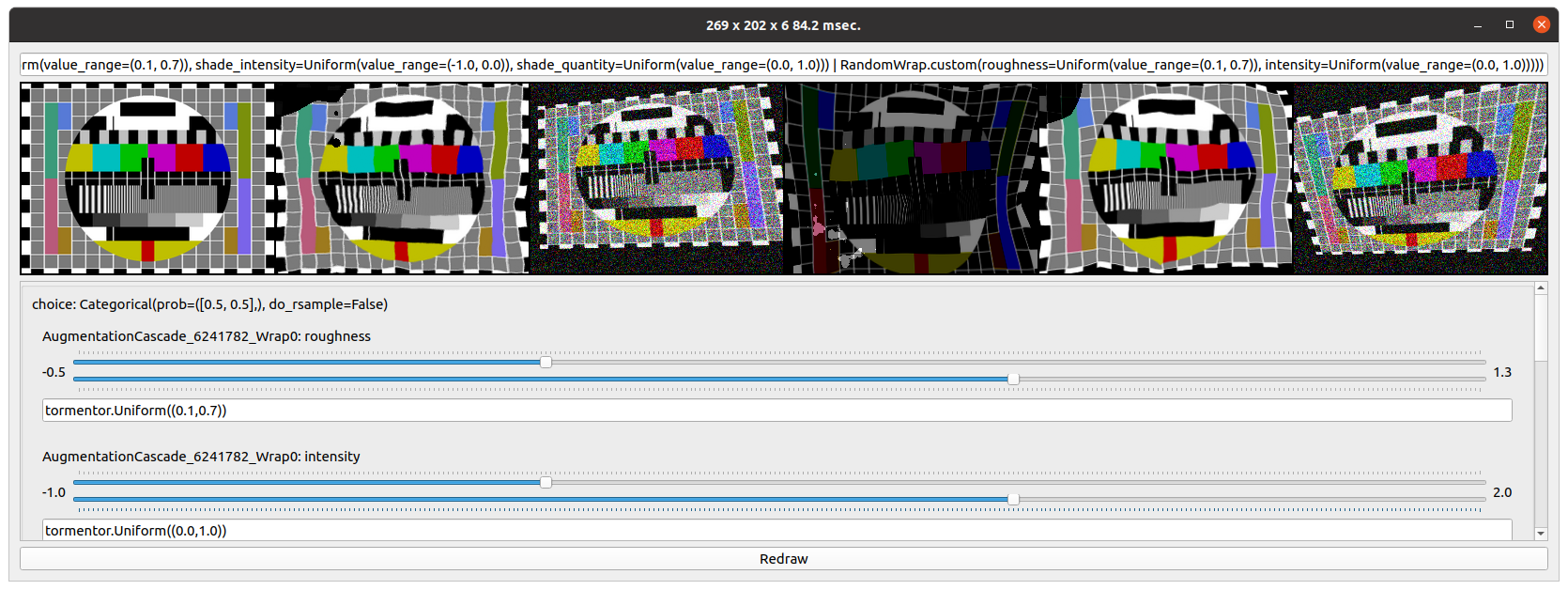}}
\caption{Tool for tuning Tormentor parameters visually}
\label{fig:tweak}
\end{figure}

In future work, we intend to replace the visual augmentation tuning tool seen in \cref{fig:tweak} with adversarial training to mimic existing data.
Extensive experiments allowing a quantitative estimation of the question ``How many fewer samples do I need to annotate if I properly tune my data augmentation?''.



\section*{Acknowledgements}
This work has been supported by the ERC Advanced Grant 101019327 'From Digital to Distant Diplomatics'. 
We would also like to thank Dmytro Mishkin and Aikaterini Symeonidi for their insights and support.
{\small
\bibliographystyle{ieee_fullname}
\bibliography{egbib}
}
\balance
\end{document}